\documentclass{article} 
\usepackage{iclr2023_conference,times}

\usepackage{amsmath,amsfonts,bm}

\def\eqref#1{equation~\ref{#1}}

\def\1{\bm{1}}

\DeclareMathAlphabet{\mathsfit}{\encodingdefault}{\sfdefault}{m}{sl}
\SetMathAlphabet{\mathsfit}{bold}{\encodingdefault}{\sfdefault}{bx}{n}

\usepackage{hyperref}
\usepackage{url}
\usepackage{graphicx}

\newcommand{\pintar}[1]{{\color{black}{#1}}}
\newcommand{\pintardos}[1]{{\color{black}{#1}}}
\title{Understanding the World to Solve Social Dilemmas using Multi-Agent Reinforcement Learning
\\
}

\author{Manuel Sebastián Ríos\thanks{We thank Google DeepMind and CINFONIA for partially funding this project through the scholarship programme.} \; \& Nicanor Quijano\\
Department of Electrical and Electronic Engineering\\
Universidad de los Andes\\
Bogotá, Colombia \\
\texttt{\{ms.rios10, nquijano\}@uniandes.edu.co} \\
\And
Luis Felipe Giraldo \\
Department of Biomedical Engineering\\
Universidad de los Andes\\
Bogotá, Colombia \\
\texttt{\{lf.giraldo404\}@uniandes.edu.co} \\
}

\iclrfinalcopy 
\begin{document}

\maketitle

\begin{abstract}
Social dilemmas are situations where groups of individuals can benefit from mutual cooperation but conflicting interests impede them from doing so. This type of situations resembles many of humanity's most critical challenges, and discovering mechanisms that facilitate the emergence of cooperative behaviors is still an open problem. In this paper, we study the behavior of self-interested rational agents that learn world models in a multi-agent reinforcement learning (RL) setting and that coexist in environments where social dilemmas can arise. Our simulation results show that groups of agents endowed with world models outperform all the other tested ones when dealing with scenarios where social dilemmas can arise. We exploit the world model architecture to qualitatively assess the learnt dynamics and confirm that each agent's world model is capable to encode information of the behavior of the changing environment and the other agent's actions. This is the first work that shows that world models \pintardos{facilitate} the emergence of complex coordinated behaviors that enable interacting agents to ``understand'' both environmental and social dynamics.

\end{abstract}

\section{Introduction}

Social dilemmas are situations where a group of individuals can benefit from the cooperativeness of its members, but they are tempted to act selfishly to satisfy their individual interests \cite{komorita2019social}. This conflict of interest is common among many of humanity's most critical challenges that include global warming, pandemic preparedness, and inequality \cite{dafoe2020open}. Understanding which mechanisms foster the emergence of cooperative behaviors that aid communities to solve social dilemmas is an important scientific question. Game theory has traditionally used matrix games to model social dilemmas. However, recent works have suggested to extend these matrix games into more dynamic and complex environments that are typically implemented in video game-like scenarios \cite{leibossd}. In this new paradigm, multi-agent RL-based algorithms have been used to model the decision-making of self-interested agents, showing how a variety of collective behaviors can emerge from these simulated environments \cite{zheng2022ai}. These behaviors have also shed light on some relevant human traits that can potentially encourage cooperation \cite{hughes2018inequity} \cite{song2022reinforcement}.

Despite that these recent works have focused their efforts on using model-free RL algorithms in multi-agent systems, theoretical analyses in social psychology have suggested that, in a group, an individual's actions are driven by his personal qualities and his own understanding of the social and changing environment they are in \cite{forsyth2018group}. 
Thus, we hypothesize that learning world models is key in multi-agent RL to study the emergent behaviors of agents that are in environments where social dilemmas arise, as world models enable each agent to ``understand'' the dynamics of a changing environment that involves social interactions. RL algorithms based on world models have gained special attention from the research community in the context of visual control and robotics \cite{ha2018recurrent} \cite{hafner2020mastering}. These model-based algorithms aim to build abstract, compact, low-dimensional representations that embed the environment's dynamics. Some authors consider that world models constitute the basis of the \textit{common sense} and are essential for building autonomous machine intelligence \cite{lecun2022path}. To our knowledge, there are no reported works that study the social dynamics of RL agents that learn world models.

We simulated the common-pool resource appropriation problem as a case of study. In this setting, a group of individuals simultaneously exploit a common resource, and it is impossible for them to exclude other individuals from using it \cite{perolat2017multi}. Generally, the resource is non-renewable or takes considerable time to renew. Therefore, exploitation decreases the available amount of resources for other individuals. A sustainable community must act in a coordinated manner, avoiding complete resource depletion. Failing to do so is commonly known as the tragedy of the commons. Our results show that world model-based RL algorithms outperform all other methods in the simulated scenarios. While the model-free algorithms fail in the task by continuously falling into the tragedy of the commons, world model-based algorithms are able to find sustainable consumption strategies. Additionally, the use of world models allows us to qualitatively assess the learnt dynamics. In this case, we show that the world model is able to encode both environmental and social dynamics. These results are consistent with theoretical models of groups in social psychology \cite{forsyth2018group} and support current trends in artificial intelligence research \cite{lecun2022path}.

\section{Related Work}

\begin{figure}
\centering
\includegraphics[width=.45\linewidth]{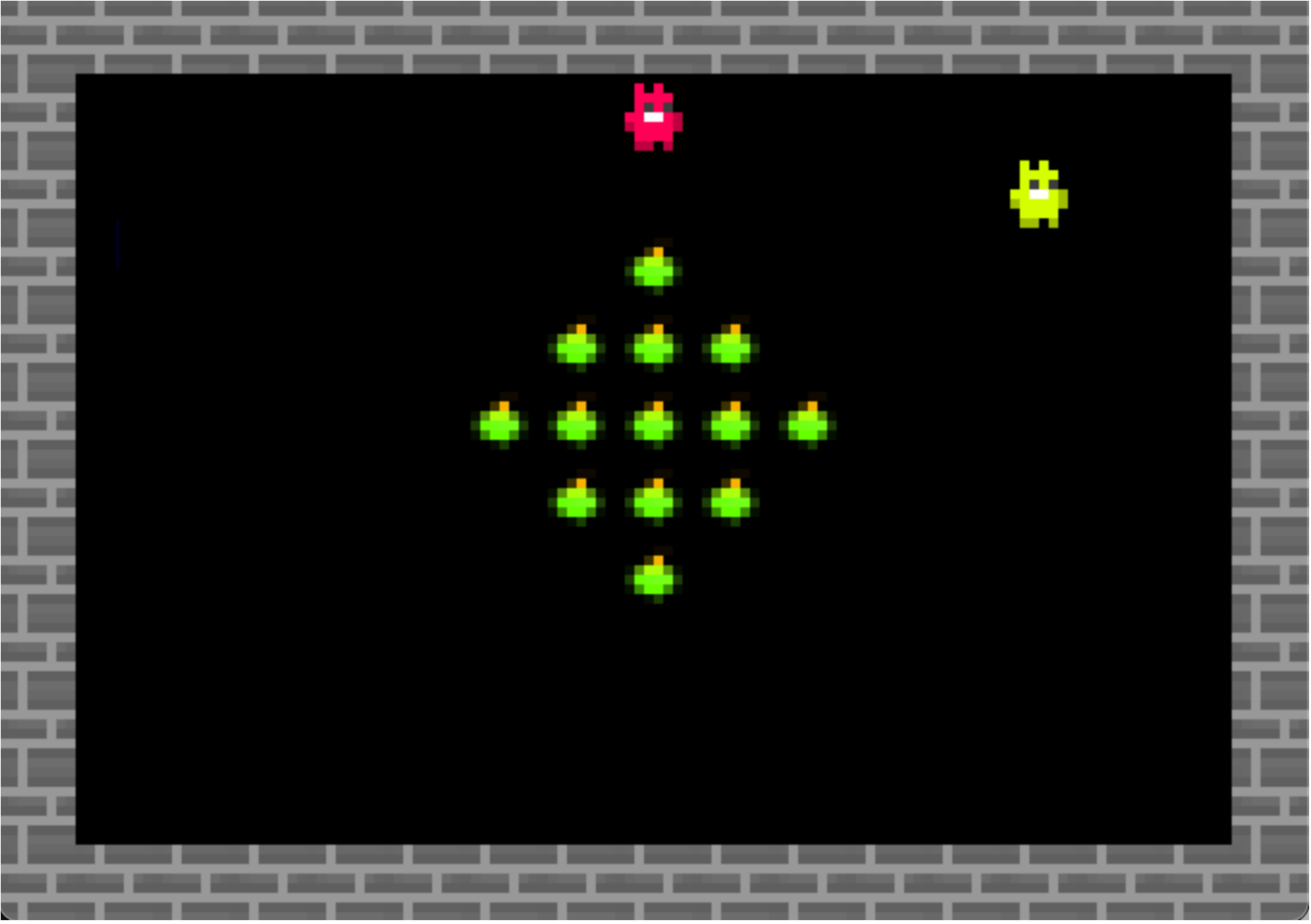}
\caption{Testing environment with two independent self-interested agents and a single apple patch.}
\label{fig:grid-world}
\end{figure}


\pintar{Social sequential dilemmas (SSDs) \cite{leibossd} can be considered as the first framework that extends social dilemmas from the classical matrix game perspective to video game-like 2D simulations. SSDs maintain the mixed motivation structure in matrix games but additionally seek to better capture some crucial aspects of real social dilemmas that include: i) social dilemmas are temporally extended; ii) cooperation and deflection are labels that should be assigned to policies instead of atomic actions; and iii) deciding whether to cooperate or deflect is done quasi-simultaneously and based only on partial information from the environment and other individuals' actions. On the other hand, finding optimal behaviors in these scenarios is computationally expensive, and requires considering the agent's high dimensional observation space. This impedes the use of classical optimization or learning techniques. Therefore, the authors propose to use reinforcement learning algorithms to train self-interested agents in these simulated scenarios. 

SSDs inspired several works that aimed to mimic some specific human traits to improve social capabilities in simulated populations using model-free reinforcement learning algorithms. For instance, \cite{hughes2018inequity} showed that modeling envy and guilt will promote the emergence of cooperative behaviors. Also, \pintardos{\cite{jaques2019social} shows that rewarding agents for having causal influence over other agents' actions promotes the emergence of coordinated behaviors, and \cite{ndousse2021emergent} show that simulated agents can benefit from the presence of expert agents to achieve complex behaviors that con not be obtained from single agent training}. These video game-like 2D environments have also been used to simulate complex economic interactions. The work in \cite{zheng2022ai} simulated a small group of self-interested agents that aimed to increase their individual income by trading, collecting, and exploiting resources. Additionally, the authors trained a planner agent to design optimal taxation policies seeking to increase the population's welfare and productivity. Remarkably, the taxation policy that emerged from this simulation outperformed many human-crafted policies.

These promising results have encouraged researchers to develop evaluation protocols and testing suits \cite{leibo2021scalable} \cite{johanson2022emergent}. Some researchers have also coined the term \textit{Cooperative AI} \cite{dafoe2020open} to the study of those mechanisms that make possible the emergence of cooperation in AI-based systems, and have made an effort to identify open problems and challenges in the field. The latter aims to guide future research in the use of AI to solve problems of cooperation in both simulated and real scenarios.}

\section{Experimental Setup}

\begin{figure}[h]
\centering
\includegraphics[width=.7\linewidth]{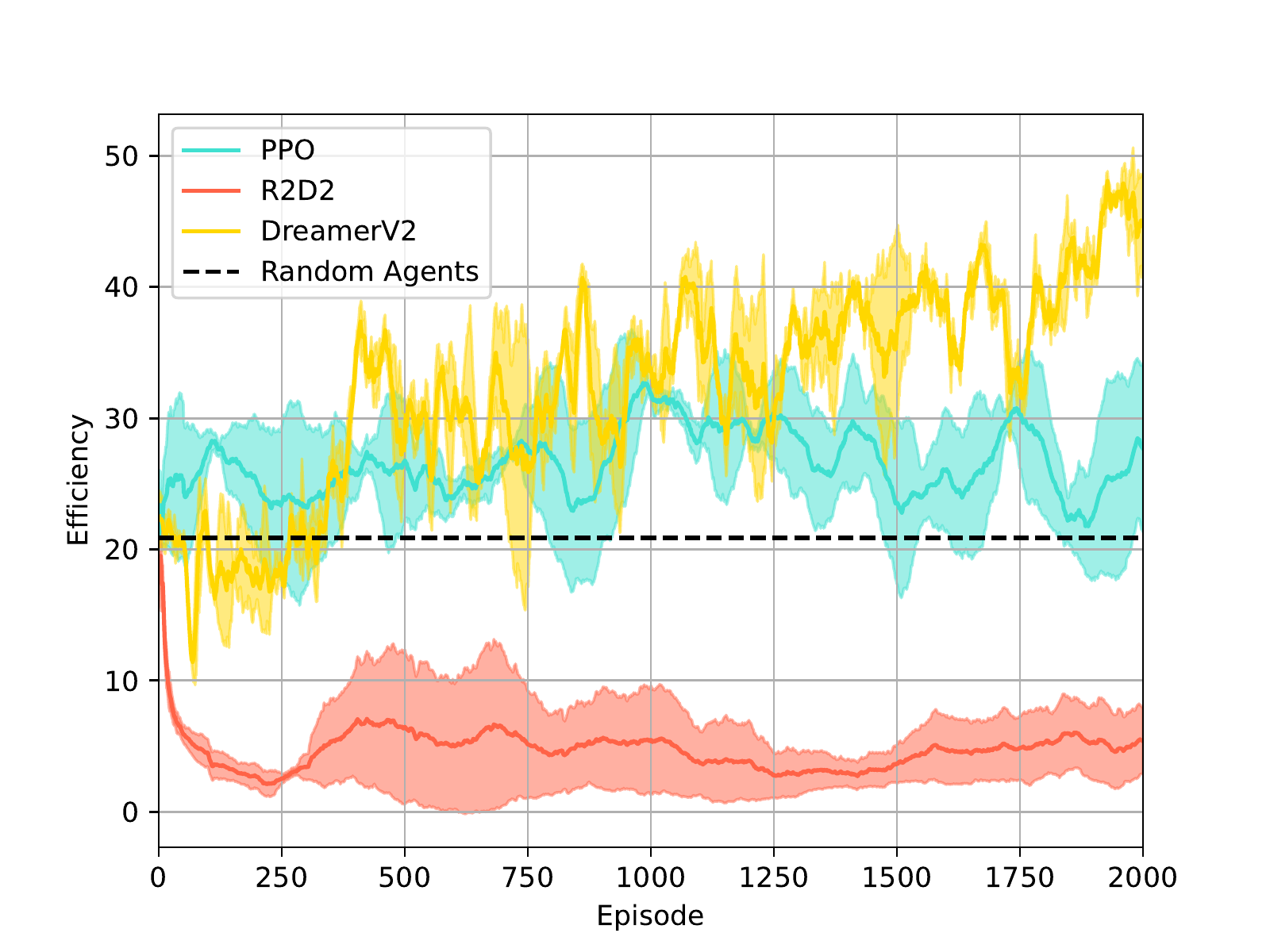}
\caption{Performance of populations trained with different algorithms on the two agent environment shown in Fig. \ref{fig:grid-world}. The performance is measured using an efficiency metric that represents the population's per-capita consumption. \pintardos{The shadowed area represents the standard deviation of the population performance over multiple experiment runs, and the solid line represents the mean performance}. The dashed line represents the expected efficiency of a population of random agents.}
\label{fig:performance}
\end{figure}

We developed our testing environment based on DeepMind's Melting Pot suit \cite{leibo2021scalable}. Inspired by Perolat, et al. \cite{perolat2017multi}, we simulated the common-pool resource appropriation problem using the grid-world shown in Fig. \ref{fig:grid-world}. In this environment, agents receive a positive reward for each apple consumed, and the apple's regrowth probability is directly proportional to the number of uneaten apples in a predefined radius. Therefore, agents must coordinate to keep at least one apple on the patch to avoid complete depletion. Agents can also use a laser beam to temporally remove other agents from the environment. Unlike previous works, our environments use a smaller regrowth probability and a lower apples-to-agents ratio to make the environment a lot more challenging to deal with. \pintar{Agents have partial observations of their environment, where they observe a small portion of the environment centered on their positions. agents must learn optimal policies directly from raw images.}

                            We trained independent agents using DreamerV2 \cite{hafner2020mastering}, which is a world model-based reinforcement learning algorithm that uses a recurrent state-space model \cite{hafner2019learning} to learn the environment's dynamics. DreamerV2 encodes these dynamics in a sparse low dimensional discrete latent state and learns optimal behavior policies by using these representations instead of the real environment's observations. We compare the performance of populations trained with DreamerV2 with populations of agents that use both off-policy (Recurrent Replay Distributed DQN R2D2) and on-policy (Proximal Policy Optimization PPO) RL algorithms. \pintardos{It is important to note that, unlike DreamerV2, both PPO and R2D2 do not use world models. To ensure a fair comparison, we trained all algorithms for 2000 episodes, following the parameters proposed by the authors in the original implementations or the parameters used in the libraries employed.} \pintar{We use the population's efficiency as a performance metric. Given a population of $N$ agents and their respective sum of rewards $R_i$, the efficiency metric can be computed as:
\begin{equation*}
    \text{Efficiency} = \frac{\sum_{i=1}^N R_i}{N}.
    \label{eq:Efficiency}
\end{equation*}

Intuitively, the efficiency represents the population's  per-capita consumption}. Fig. \ref{fig:performance} depicts the performance of all the tested populations. Our results clearly show how the agents trained using DreamerV2 outperform both PPO and R2D2. We provide videos of the learnt policies for all the trained agents (\href{https://youtu.be/5qkPw4m9v9o}{supporting video}) and all the source code used to conduct this research \footnote{\href{https://github.com/ManuelRios18/Commons-Tragedy}{https://github.com/ManuelRios18/Commons-Tragedy}}.

\begin{figure*}[h!]
\centering
\includegraphics[scale=0.45]{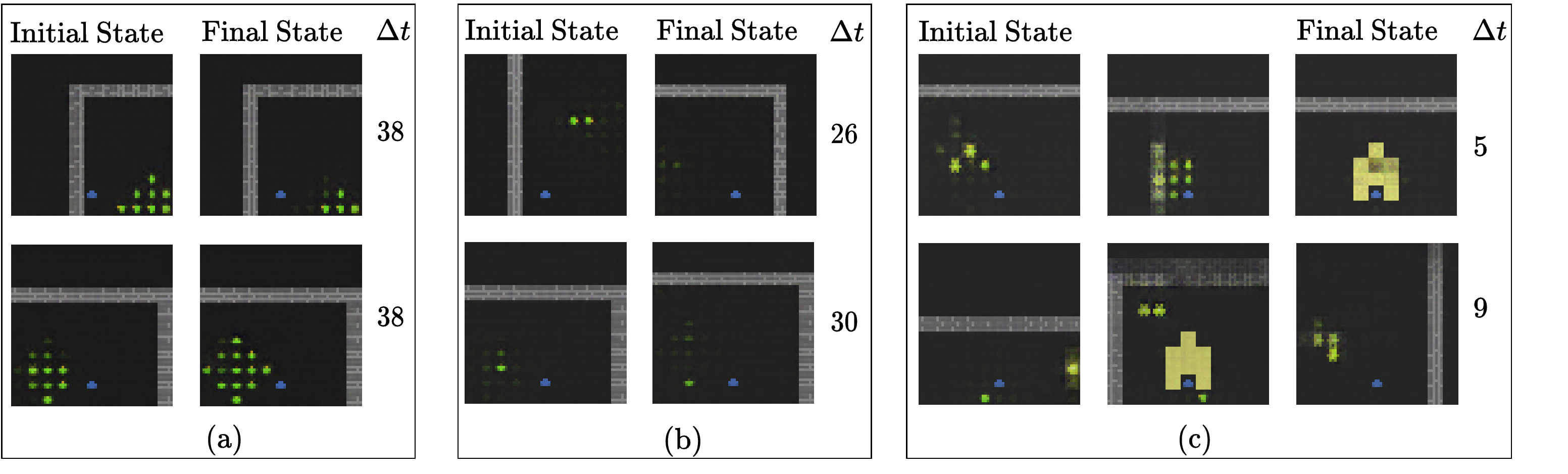}
\caption{Initial and final states sampled from predicted trajectories when: \textbf{(a)} there is a high density of uneaten apples in the initial states; \textbf{(b)} uneaten apples are scarce in the initial states; \textbf{(c)} there are interactions with the other agent (the yellow shaded area in front of the agent is the laser beam). $\Delta t$ denotes the number of states between the initial and final states.}
\label{fig:dreams}
\end{figure*}

One of the key features of DreamerV2 is that each agent after the learning process is able to predict hypothetical future state sequences from a single initial observed state. That is, the world model. Therefore, it is possible to qualitatively assess the agent's learnt behaviors by predicting state sequences from key initial observations. Fig. \ref{fig:dreams} shows the final predicted states computed from initial states with different densities of uneaten apples and the presence of other agents. These results suggest that the algorithm is able to properly encode the environment's dynamics. The apples' density in the final state is directly proportional to the density in the initial state. Additionally, DreamerV2 is also able to model other agents' behavior by predicting their intentions of consuming the apples. Moreover, the model also predicts future attacks to the other agents, proving that the model understands that it is beneficial to reduce the effective population, given that this allows the agent to consume the apples without taking the risk of being taken out of the environment.


\begin{figure*}[h!]
\centering
\includegraphics[scale=0.50]{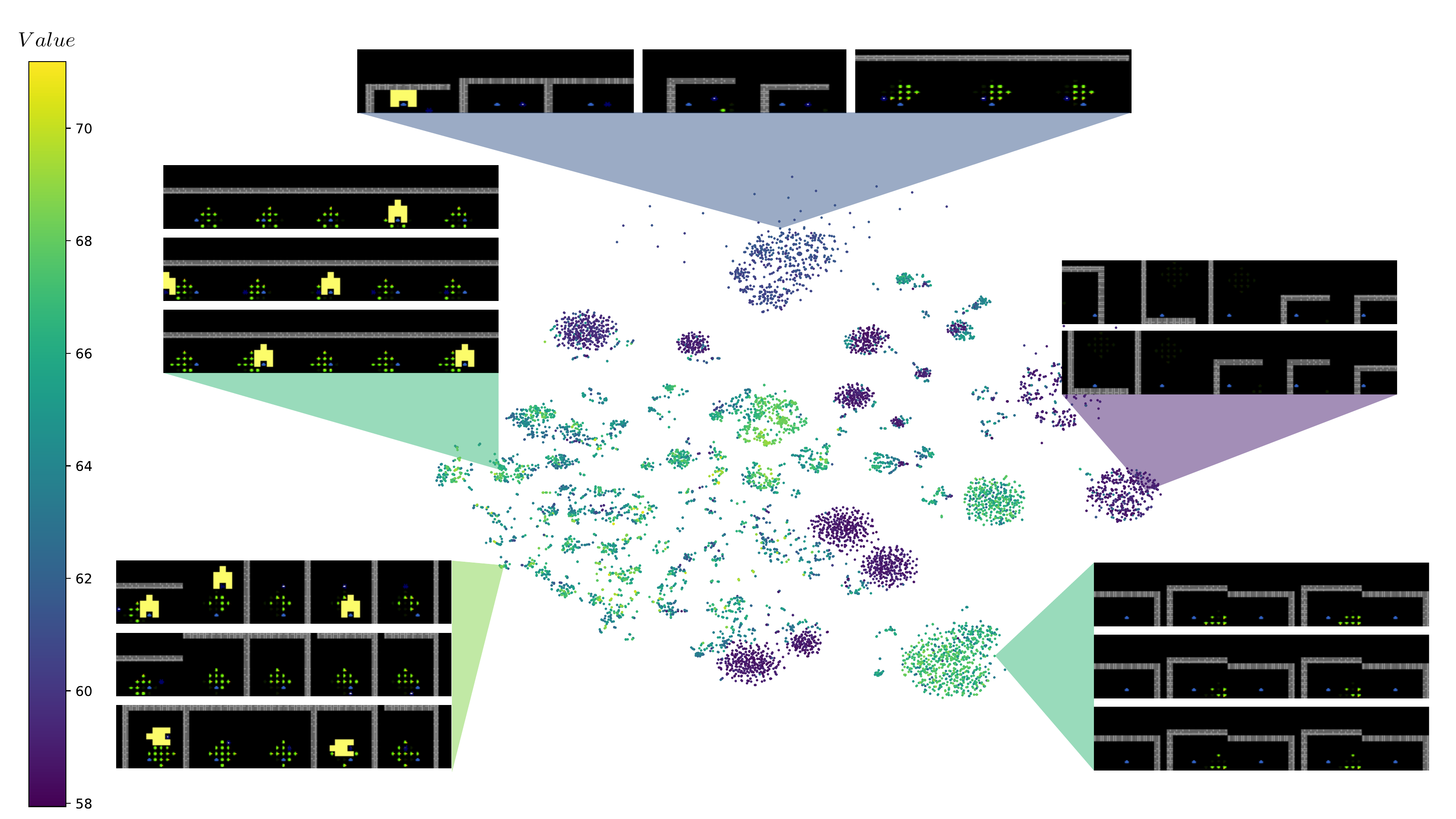}
\caption{t-SNE projection of different latent states, and the observation sequence used to compute some of these latent states. Each state is colored according to its corresponding state value.}
\label{fig:tsne}
\end{figure*}

Fig. \ref{fig:tsne} shows the t-SNE projection of many latent states obtained with a trained agent. The colors represent the state value, which can be interpreted as a scalar number that indicates how good the agent is to be in a given state in terms of the expected sum of rewards. Our results show that the learnt model groups together states that share similar state values and also environmental and social dynamics. For instance, the cluster located in the upper middle section of Fig. \ref{fig:tsne} is composed of observation sequences where the agents interacted with each other. The risk of being too close to the other agent, the low apple density, and receiving a direct attack may explain the low value assigned to this set of latent states.

In the case of the common-pool appropriation problem, understanding the environmental dynamics helps the agents to avoid eating the last apple in the patch. As shown in our supporting videos, when there is a single apple left, the agents patiently wait for the other apples to grow back and never completely deplete the patch. Fig. \ref{fig:dreams} shows that DreamerV2 excels at modeling other agents' intentions. The predicted trajectories show that the other agents are perceived as a threat that must be attacked. Moreover, some predicted states show these agents in the middle of the map consuming the apples. Generally, the other agent is represented as a blurry patch over many adjacent cells showing that the model captures the uncertainty associated with the other's actions. Additionally, our t-SNE projection shows that the states are clustered based on both the presence of the other agents and the current apples' density. This suggests that both social and environmental dynamics shape the learnt policies.

\section{Discussion}

The results presented in this study suggest that world models can considerably ease the emergence of coordinated behaviors in self-interested individuals. The fact that the world models encode social and environmental dynamics and that the agents exploit this information to compute sustainable behavior policies, is consistent with theoretical models in social psychology and current artificial intelligence research directions \cite{lecun2022path}. Understanding both the environmental dynamics and the intentions of the other agents is considered one of the key elements in cooperative intelligence \cite{dafoe2020open} and was crucial for the emergence of these complex coordinated behaviors. 

Also, we highlight the use of discrete representations to encode the world's dynamics. \cite{ma2022principles} argue that \textit{parsimony} is a cornerstone for the emergence of intelligence, and \cite{gomez2020learning} showed the benefits of using discrete representations to facilitate the ability to generalize to novel situations. In this case, using discrete and sparse arrays as latent states of the world models ensures compactness and simplicity, potentially allowing agents to model more challenging social scenarios. 

We consider that this is a promising research direction that can lead to intelligent systems to aid decision-makers in complex real-world challenges that involve social dynamics.

\bibliography{iclr2023_conference}
\bibliographystyle{iclr2023_conference}


\end{document}